# Direct off-line robot programming via a common CAD package

Pedro Neto, Nuno Mendes

Department of Mechanical Engineering – POLO II, University of Coimbra, 3030-708 Coimbra, Portugal.

**Corresponding author:** Pedro Neto email: pedro.neto@dem.uc.pt; tel: +351 239 790 700

*Abstract*:

This paper focuses on intuitive and direct off-line robot programming from a CAD drawing running on a common 3-D CAD package, *Autodesk Inventor*. It explores the most suitable way to represent robot motion in a CAD drawing, how to automatically extract that motion data from the drawing, mapping it from the virtual (CAD model) to the real scenario and the process of automatic generation of robot programs. This study aims to create a CAD-based robot programming system accessible to anyone with basic knowledge of CAD and robotics and thus contribute to increase the number of existing robots. Experiments on different manipulation tasks show the effectiveness and versatility of the proposed approach.

## 1. Introduction

Robot programming through the conventional teaching process (using the teach pendant) is often a tedious and time-consuming task that demands significant technical expertise. Many times, companies do not have robots and/or other automatic systems in their facilities because they know that the configuration and programming process of this type of equipments is time-consuming and requires workers with knowledge in the field. Nevertheless, most industrial robots are still programmed using the conventional teaching process. In this way, new and more intuitive approaches to robot programming are required.

Drawing inspiration from the way humans communicate with each other, this paper explores and studies methodologies that can help robot users to interact with a robot in an intuitive way, with a high-level of abstraction from the robot specific language. In fact, a human being can be taught in several different ways, for example, through drawings. As an example, it is very common to see a human being explaining something to another human being with base on a CAD drawing.

In this paper, a new CAD-based off-line robot programming (OLP) system is proposed. Robot programs are directly generated from a 3-D CAD drawing running on a commonly available 3-D CAD package (*Autodesk Inventor*) and not from commercial OLP or CAM software. The aim is to automatically generate robot motion sequences from a graphical description of the robot paths over a 3-D

CAD model of a given robotic cell. A unified treatment of CAD and robot programming methods may involve very important advances in versatility and autonomy of the platform, in other words, product design and robot programming can be integrated seamlessly. It is explored the most suitable way to a user represent robot motion in a CAD drawing, how to automatically extract it from drawings, make the mapping between the virtual and the real scenario and automatically generate robot programs.

This work aims to create a CAD-based OLP system accessible to anyone with basic knowledge of CAD and robotics. This can open the door to new robot users and thus contribute to increase the number of existing robots in companies. Some algorithms with running code are presented, allowing readers to replicate and improve the work done so far. Experiments on different manipulation tasks show the effectiveness and versatility of the proposed approach.

## 2. Off-line robot programming

Contrary to on-line programming, in OLP robots are programmed without stopping/disturbing robot production, Fig. 1. Robot programs are created in a computer, without the need to access the robot [1]. This may involve manual editing of robot code and/or the definition of the robot programs by means of computer software that simulates the real robotic scenario. Some major advantages of OLP include:

- Robot programming without stopping/disturbing production. Robots can be programmed before installation and stay in production while being re-programmed for a new task. This means that robot programming can be carried out in parallel with production (production breaks are shortened);
- The programming efforts are moved from the robot operator in the workshop to the engineer in the office;
- Increase of work safety. During the programming process the user is not in the robot working area;
- Robot programs can be tested using simulation tools. This is very important to anticipate/predict the real robots behaviour and in this way to optimize working processes.

On the other hand, some disadvantages can be pointed out:
- Relatively high initial investment in software and workers' training. This investment is difficult to justify for most small and medium-sized enterprises (SMEs);
- The calibration process requires experienced operators. A rough calibration process can lead to tremendous inaccuracies during robot operation;
- Process information is required in advance;
- OLP methods rely on accurate modelling of the robotic cell.

**Fig. 1.** OLP concept.

Software packages dedicated to OLP are usually called OLP software or computer-aided robotics (CAR) software. Some OLP packages are able to operate with robots from different manufacturers (generic OLP packages). Three of the most common generic OLP packages are *Delmia* from *Dassault*

*Systèmes*, *RobCAD* from *Technomatix Technologies* and *Robotmaster* from *Jabez Technologies*. These software packages provide a set of modelling and simulation tools capable to represent graphically a robot manipulator and its attendant equipment, generate programs, and hence simulate a given manufacturing task [2]-[3]. On the other hand, almost every robot manufacturer has its own OLP software. Examples are *RobotStudio* from *ABB Robotics* and *MotoSim* from *Motoman*. Early versions of OLP software were based on simple wireframe models of the robot's kinematics. However, in recent years, robot simulation techniques have seen a rise in realism and popularity, possibly coinciding with the advancement of computing and graphical animation technologies. OLP packages of today are more graphically powerful, modular (with modules for specific processes such as coating and welding) and standard (with capacity for example to import standard CAD formats).

All of these capabilities come at a cost. A license for OLP software can costs thousands of U.S. Dollars, an investment difficult to justify for most SMEs. Advantages of OLP software are tempered by some limitations in existing systems. In fact, they are not intuitive to use and can only be applied in situations where the robot surrounding environment is known a priori and well modelled [4]. In addition, the calibration process associated to OLP continues to be a laborious task and likely to create error.

*2.1. CAD-based robot programming*

In recent years, CAD technology has become economically attractive and easy to work with. Today, millions of SMEs worldwide are using CAD technology to design and model their products. Nevertheless, the CAD industry has to face significant technical challenges in future [5].

Already in the 80's, CAD was seen as a technology that could help in the development of robotics [6]. Since then, a variety of research has been conducted in the field of CAD-based robot planning and programming. Over the years some researchers have explored CAD technology trying to extend its capabilities to the robotics field. Today, it is possible to extract information from CAD drawings/files to generate robot programs for many different applications. For example, robot paths can be extracted from CAD [7]-[10].

A series of studies have been conducted using CAD as an interface between robots and humans. Diverse solutions have been proposed for the processes of spray painting and coating. A review of CAD-based robot path planning for spray painting is presented by Chen *et al*. [11]. A CAD-guided robot path generator is proposed for the process of spray painting of compound surfaces commonly seen in automotive manufacturing [12]. Arikan and Balkan [13] propose a CAD-based robotic system addressing the spray painting process of curved surfaces (OLP and simulation) and Chen *et al*. [14] a CAD-based automated robot trajectory planning system for spray painting of free-form surfaces.

An important study in the field of CAD-based robotics presents a method to generate 3-D robot working paths for a robotic adhesive spray system for shoe outsoles and uppers [15]. An example of a novel process that benefits from robots and CAD versatility is the so-called incremental forming process of metal sheets. Without using any costly form, metal sheets are clamped in a rigid frame and the robot produces a given 3-D contour by guiding a tool equipped with a high-frequency oscillating stamp over the metal surface. The robot's trajectories are computed from the CAD model on the basis of specific material models. Prototype panels or customized car panels can be economically produced using this method [16]. Pulkkinen *et al.* [17] present a robot programming concept for applications where metal profiles are processed by robots and only a 2-D geometrical representation of the workpiece is available.

Nagata *et al*. [18] propose a robotic sanding platform where robot paths are generated by CAD/CAM software. A recent study discusses robot path generation from CAM software for rapid prototyping applications [19]. Feng-yun and Tian-sheng [20] present a robot path generator for a polishing process where cutter location data is generated from the postprocessor of a CAD system. Other previous studies report the development of robotic systems for rapid prototyping in which cutting data are extracted from CAD drawings [21]-[22]. A CAD-based system to generate deburring paths from CAD data is proposed by Murphy *et al.* [23]. A method for manufacturing prototype castings using a robot-based system in which manufacturing paths are generated from a CAM system is proposed by Sallinen and Servio [24]. In a different kind of application, CAD drawings are used for robot navigation purposes, namely for large scale path planning [25].

As we have seen above, a variety of research has been conducted in the fields of CAD-, CAM- and VRML-based OLP. However, none of the studies so far has an effective solution for an intuitive and low-cost OLP solution using raw CAD data and directly interfacing with a commercial CAD package. Research studies in this area have produced great results, some of them already implemented in industry, but limited to a specific industrial process (welding, painting, etc.). Even though a variety of approaches has been presented, a cost-effective standard solution has not been established yet.

## 3. CAD-based approach

### 3.1. CAD packages

CAD technology has become economically attractive and easy to work with so that today there are millions of companies worldwide using it to design and model their products. While the prices of CAD packages have decreased, their features and functionalities have been upgraded, with improved and simplified user interfaces, user-oriented functionalities, automatic design of standard products, etc. Nowadays, most CAD packages provide a wide range of associated features (integrated modules or standalone solutions) that not only help in the effective design process, but also help in other tasks such as mechanical simulation and the physical simulation of dynamic processes. Robot programming and simulation has been seen as another feature that CAD packages can integrate.

#### 3.1.1. Autodesk Inventor

*Autodesk Inventor* is one of the most common 3-D CAD packages of today. It incorporates all the functionalities of modern CAD packages:
- High-level of sophistication in terms of CAD technology, design, visualization and simulation;
- User-friendly interface;
- It provides a complete application programming interface (API) for customization purposes, allowing developers to customize their CAD-based applications [26];
- Strong presence in the market.

In terms of file formats, besides all the standard formats, *Autodesk Inventor* has proprietary file formats to define:

- *Parts* (ipt file). File format for single *part model* files;
- *Assemblies* (iam file). File format for *assembly model* files.

*3.2. Extracting data from CAD*

The base of the proposed CAD-based OLP platform is the ability to automatically extract robot motion from CAD drawings running on *Autodesk Inventor*. The *Autodesk Inventor API* is used for that purpose. It exposes the *Inventor's* functionalities in an object-oriented manner using a technology from *Microsoft* called *Automation*. In this way, developers can interact with *Autodesk Inventor* using current programming languages such as Visual Basic (VB), Visual C# and Visual C++. The *API* allows developers to create a software interface that performs the same type of operations that a user can perform when using *Autodesk Inventor* interactively. Summarizing, the *Autodesk Inventor API* provides a set of routines that may be used to build a software interface based on resources from *Autodesk Inventor*.

There are different ways to access the *Autodesk Inventor API* [27]. Fig. 2 shows all the possible ways to access the *Autodesk Inventor API*. The white boxes represent components provided by the *API* (*Autodesk Inventor* and *Apprentice Server*) and the soft blue boxes represent programs written by developers. When one box encloses another box, this indicates that the enclosed box is running in the same process as the box which is enclosing it. In this way, an "in-process" program will run significantly faster than a program running out of the process.

**Fig. 2.** Accessing the Autodesk Inventor's API.

All the different ways to access the *API* are useful in specific cases, so that it is important to choose the most appropriate way to access it. Considering the solution presented in this paper, a standalone application is proposed to access the *API* and subsequently *Autodesk Inventor* data. This choice was due to the necessity to integrate in the main application not only the process of interaction with CAD but also other software components for other tasks, for example, robot communications.

3.2.1. Using the Autodesk Inventor API

The *API* provided by *Autodesk* is a very well structured and documented API in such a way that it is relatively simple to access information contained in a CAD drawing. As an example, it is possible through a standalone application to open *Autodesk Inventor* in visible mode, Fig. 3, and open an *Inventor* document, Fig. 4. The properties of the open document can then be easily accessed, including the name of the document, Fig. 4.

**Fig. 3.** Opening *Autodesk Inventor* (coded in VB).

**Fig. 4.** Opening an *Inventor* document and extracting its properties and name (coded in VB).

3.2.2. Extracted data

There are a great number of data that can be extracted from a CAD drawing. First, it is necessary to establish what type of data we need. We need to extract robot motion from CAD drawings, a sequence of target points that represent the robot end-effector poses (position and orientation) with respect to a known coordinate system (in Cartesian space). Thus, given the capacity of the *Autodesk Inventor API* (for good and for evil), it was established that we need to extract positions and orientations of objects in 3-D space from proper CAD drawings representing the robotic cell in study:

1. Positions. Positional data are acquired from a CAD drawing through different ways, for example, acquiring *WorkPoints* positional data (Fig. 5). In other situations, positional data come from the points that characterize each one of the different lines representing virtual robot paths in a CAD drawing, Fig. 6. All these data are defined in relation to the origin of the CAD *assembly model* of the robotic cell;

2. Orientations. The *API* provides information about the transformation matrix (or homogeneous transform) of each *part model* represented in a CAD *assembly model*, Fig. 7. The transformation matrix contains the rotation matrix and the position of the origin of the *part model* to which it refers, both in relation to the origin of the CAD *assembly model* of the robotic cell.

**Fig. 5.** Extracting data from a selected *WorkPoint* (coded in VB).

**Fig. 6.** Extracting data from a selected virtual line (coded in VB).

**Fig. 7.** Extracting the transformation matrix of a selected item (coded in VB).

*3.3. CAD models*

The process of creating the CAD *part models* that compose the CAD *assembly model* of the robotic cell should respect some rules. Since it was established in section 3.2.2. that we need to have represented in the CAD *assembly model* of the robotic cell all the required robot paths (end-effector poses), it becomes necessary to study the most suitable way to have that information represented in CAD models. This can be achieved in two different ways:

1. By introducing extra robot tool (end-effector) models within the *assembly model*. These models should represent the desired robot end-effector pose in each segment of the path, Fig. 8.

Positional data are achieved by placing a *WorkPoint* attached to any part of the tool model, Fig. 9. The *WorkPoints* data are provided by the *API* in relation to the origin of the CAD *assembly model*. Orientation data are achieved from the tool models orientation in the drawing;

2. By drawing lines (in the *assembly model*) representing the desired robot path (positional data). Furthermore, we need to define the robot end-effector orientation in space. Therefore, after drawing the robot paths, simplified tool models should be placed along such path lines to achieve the desired end-effector orientations, as in the above topic, Fig. 8.

**Fig. 8.** Simplified tool models defining the robot end-effector pose in each path segment.

**Fig. 9.** A *WorkPoint* attached to a tool model in a location where the tool is connected to the robot wrist.

The CAD *assembly model* does not need to accurately represent the real cell in all its aspects. On the contrary, it can be a simplified model containing all important information. As an example, the robot tool length, robot paths and relative positioning of CAD models should accurately represent the real scenario. However, the models appearance does not need to be exactly the same as the real objects. It means that, for example, chamfers or rounded edges are expendable. These simplifications allow to speed up the modelling process. Fig. 10 shows a real robot tool (a) and two CAD models of that tool, (b) and (c), with the same length $l$. These two models were created with different levels of detail:

1. Model (b) was created with more detail than model (c). Model (b) represents more accurately the real tool, providing in this way some advantages in terms of visualization. Nevertheless, the process of drawing this model is more time consuming than drawing model (c);

2. Model (c) is a simpler version but accurate at the same time in terms of length. It can be drawn in seconds and used where only the length of the tool is a factor of importance.

At this stage the designer has to make a choice about the level of detail of each model. It is important to note that the best model is the simplest model that still serves its purpose.

**Fig. 10.** Real robot tool (a), and simplified models (b) and (c).

3.3.1. Process planning

In order to achieve feasible robot paths, the user has to plan ahead some process issues and resort to them during the definition of the CAD *assembly model*. In fact, during the definition/construction of the CAD *assembly model* the user is planning in advance the "best" process parameters and paths. Depending on the type and complexity of the process in study, the planning task can include several factors:

1. Models selection/construction and definition of the layout of the cell. Some CAD models can be accessed from libraries provided by manufacturers (robots and other peripheral equipment). Other cell components have to be designed by the users. When the layout of the cell is changed (this includes small adjustments) the CAD model of the cell must accompany this change. CAD models can be reused to create an updated layout;

2. Robot motion. Robot motion is indirectly defined by placing simplified tool (end-effector) models and virtual paths within the CAD *assembly model*. The desired movement type (linear, circular, joint or spline) is defined by the virtual paths above mentioned or in the software interface. Robot velocity is also defined in the software interface by the user. The designer should plan the layout of the cell appropriately, ensuring that the real robot reaches all the planned working spaces;

3. Operation sequences. Robot operation sequences are defined by the designer by naming the tool models with a specific name. The first five characters of the name of a tool model should be "step_". The sixth character and following should be a number defining the ordering sequence. Following the sequence number the tool name can have or not a letter indicating a specific robot operation, for example, "step_1A" can indicate a robot motion plus the activation of a digital output of the robot (this one can be associated to a grasping operation);

4. Collisions. Collisions should be predicted by the designer during the creation of the CAD model of the cell. The designer should ensure that there are no collisions between the robot and other objects within the workspace. Fig. 11 shows a CAD drawing with two tool models (initial and target pose) and an obstacle. If the robot end-effector is linearly moved from the initial to the target pose a collision occurs. The designer should anticipate this situation and introduce into the drawing "intermediate" tool models to allow the robot to avoid the obstacle, Fig. 12;

5. Grasping and re-grasping/repositioning. These are common situations in industrial robotics, especially in pure manipulation tasks. Many times, in order to properly perform a task, there is a need for re-grasping or repositioning of a given workpiece. Fig. 13 shows a re-grasping process in which a grasping location is changed from an initial pose defined by the tool model step_1 to a target pose defined by step_5. Moreover, during the planning phase, the designer should ensure that the robot is operating with valid tool/grip locations, including good contact conditions between the gripper and the workpiece. Note that as in Fig. 12, also in Fig. 13 some "intermediate" tool models are used to avoid collisions during the re-grasping process.

A robot simulation system can be a valuable help in this planning phase, helping users to visualize the robotic process (robot motion, possible collisions and the re-grasping operations) and detect existing robot kinematic singularities or robot joint limits [28].

**Fig. 11.** Collision.

**Fig. 12.** Avoiding an obstacle.

**Fig. 13.** Re-grasping.

*3.4. Mapping and calibration*

Many times we need to express the same quantity in relation to different coordinate systems, in other words, change descriptions from frame to frame. This is usually called mapping. These capabilities can be used for the calibration process, making the CAD model of the cell in study to match with the real robotic cell. All robot end-effector positions and orientations extracted from CAD have to be known with respect to one or more reference frames known a priori by the real robot. These frames are made known to the robot through a calibration process specifically created for this purpose. The frame(s) have to be defined within the CAD drawing of the cell. This can be achieved by placing an invisible *part model* with the desired frame pose into the CAD *assembly model* (note that each *part model* has a frame associated). Then, the real robot is taught about that frame(s)' pose in the real scenario through the conventional way, using the teach pendant. Essentially, the process consists in the definition of one or more frames within the CAD drawing of the cell and in the robot controller. This makes calibration a relatively simple and non-time consuming process. Nevertheless, more complex robotic scenarios can require the definition of a significant number of different frames. In this case the calibration process can be lengthy and prone to error. This is because the user has to remember the pose of each frame previously defined within the CAD drawing and at the same time to define such frames in the real scenario.

As mentioned before, the *Autodesk Inventor API* provides all the information (transformation matrices, *WorkPoints* and path lines data) with respect to the origin of the CAD *assembly model*, here defined by frame {U}, Fig. 14. Frame {B} is defined in {R} during the calibration process (in the real robot), and at the same time the *API* provides the transformation matrix of {B} relative to {U}, ${}^{U}_{B}\mathbf{T}$. This means that frame {B} "makes the link" between the virtual and real world. Note that, as mentioned above, it is possible to define more than one frame if necessary, as the process is similar.

Since *Autodesk Inventor* considers the tool models (with a *WorkPoint* attached) and the path lines as a constituent of a single *part model* contained in the CAD *assembly model*, the transformation matrix (relative to {U}) of that single *part model* defines the pose of tool models and path lines. For the general case presented in Fig. 14 the path line is part of the table top model. The table top model has the origin and orientation defined by {E}. However, it is not necessary to know the orientation of the path lines because the *API* gives all the necessary points to define the path lines relative to {U}, for example the initial path point ${}^{U}\mathbf{P}_{ini}$, Fig. 14. In this way, it is necessary to achieve the path line points relative to frame {B}. The same for the tool models in which we need to have orientations and *WorkPoint* positional data relative to {B}.

The generic tool models that incorporate {C} and {D}, Fig. 14, help to define the end-effector pose in each path segment, as well as the *WorkPoint* positions (if they have a *WorkPoint* attached). The *API* provides the transformation matrix of these models relative to {U}, ${}^{U}_{C}\mathbf{T}$ and ${}^{U}_{D}\mathbf{T}$. Given our purpose

(robot programming), we wish to express {C} and {D} in terms of {B}, $_C^B\mathbf{T}$ and $_D^B\mathbf{T}$. For $_C^B\mathbf{T}$ we have that:

$$_C^B\mathbf{T} = {}_U^B\mathbf{T}\,{}_C^U\mathbf{T} \tag{1}$$

To find $_U^B\mathbf{T}$, we must compute the rotation matrix that defines frame {U} relative to {B}, $_U^B\mathbf{R}$, and the vector that locates the origin of frame {U} relative to {B}, $^B\mathbf{P}_{Uorg}$:

$$_U^B\mathbf{T} = \begin{bmatrix} _U^B\mathbf{R} & ^B\mathbf{P}_{Uorg} \\ 0\ 0\ 0 & 1 \end{bmatrix} \tag{2}$$

Let's consider a generic vector/point defined in {U}, $^U\mathbf{P}$. If we wish to express this point in space in terms of frame {B} we must compute:

$$^B\mathbf{P} = {}_U^B\mathbf{R}\,{}^U\mathbf{P} + {}^B\mathbf{P}_{Uorg} \tag{3}$$

Given the characteristics of a rotation matrix, $_U^B\mathbf{R} = {}_B^U\mathbf{R}^T$, and as we know $_B^U\mathbf{T}$, there follows the computation of $^B\mathbf{P}_{Uorg}$. From the process of inverting a transform we have that:

$$_U^B\mathbf{T} = \begin{bmatrix} _B^U\mathbf{R}^T & -{}_B^U\mathbf{R}^T\,{}^U\mathbf{P}_{Borg} \\ 0\ 0\ 0 & 1 \end{bmatrix} \tag{4}$$

In this way, from (2) and (4) we have that:

$$^B\mathbf{P}_{Uorg} = -{}_B^U\mathbf{R}^T\,{}^U\mathbf{P}_{Borg} \tag{5}$$

Now, from (1) and (4) we can compute $_C^B\mathbf{T}$. The same methodology can be applied to achieve $_D^B\mathbf{T}$ and any other transformation. This means that all positions and orientations extracted from CAD can be referred with respect to a single or more reference frame(s) defined by the user at the moment of the calibration process.

**Fig. 14.** Coordinate frames.

*3.5. X-Y-Z Euler angles*

After having obtained from CAD the rotation matrices defining robot end-effector orientations in relation to a given frame, it becomes necessary to transform such matrices into effective end-effector rotations, usually Euler angles or quaternions.

The description of the orientation of a frame {B} with respect to a frame {A} in the form of X-Y-Z Euler angles ($\alpha$, $\beta$ and $\gamma$) can be represented by a rotation matrix:

$$_{B}^{A}\mathbf{Rot}_{xyz} = \mathbf{Rot}_{x}(\alpha)\,\mathbf{Rot}_{y}(\beta)\,\mathbf{Rot}_{z}(\gamma) \tag{6}$$

In matrix form:

$$_{B}^{A}\mathbf{Rot}_{xyz} = \begin{bmatrix} r_{1,1} & r_{1,2} & r_{1,3} \\ r_{2,1} & r_{2,2} & r_{2,3} \\ r_{3,1} & r_{3,2} & r_{3,3} \end{bmatrix} \tag{7}$$

It is now possible to compute the X-Y-Z Euler angles. If $\beta \neq \pm(\pi/2)$:

$$\beta = A\tan 2\left(r_{1,3}, \sqrt{r_{1,1}^2 + r_{1,2}^2}\right) \tag{8}$$

$$\alpha = A\tan 2\left(\frac{r_{2,3}}{-\sqrt{r_{1,1}^2 + r_{1,2}^2}}, \frac{r_{3,3}}{\sqrt{r_{1,1}^2 + r_{1,2}^2}}\right) \tag{9}$$

$$\gamma = A\tan 2\left(\frac{r_{1,2}}{-\sqrt{r_{1,1}^2 + r_{1,2}^2}}, \frac{r_{1,1}}{\sqrt{r_{1,1}^2 + r_{1,2}^2}}\right) \tag{10}$$

Where Atan2(*y*,*x*) is a two argument arc tangent function. When $\beta = \pm(\pi/2)$, the process to achieve Euler angles is more complicated. In this situation we have that both the *x* and *z* axes are aligned with each other and one degree of freedom is lost. This phenomenon is mathematically unsolvable and is known as gimbal lock. In this scenario, $\alpha$ and $\gamma$ cannot be calculated separately but together:

$$\alpha \pm \gamma = A\tan 2\left(r_{3,2}, r_{2,2}\right) \tag{11}$$

The gimbal lock phenomenon does not make Euler angles "wrong" but makes them unsuited for some practical applications. Some methods have been proposed to deal with the gimbal lock phenomenon, for example, solutions based on the representation of rigid body orientation through quaternions [29]. However, some robot manufacturers force the use of Euler angles so that the option for quaternions is ruled out. In a different approach to try to solve the problem, Pollard *et al.* [30] propose to locate regions near gimbal lock and compute a restricted degree of freedom solution within those regions. Nevertheless, in practice, a typical approach (convention) is to set an angle equal to zero and compute the remaining angle. In this case, if $\beta = (\pi/2)$, and assuming that $\alpha = 0$, we have:

$$\gamma = A\tan 2\left(r_{2,1}, -r_{3,1}\right) \tag{12}$$

On contrary, if $\beta = -(\pi/2)$ and assuming that $\alpha = 0$, we have:

$$\gamma = A\tan 2(r_{2,1}, r_{3,1}) \tag{13}$$

*3.6. Interpolation*

When an industrial robot is performing a pre-programmed movement and this one requires abrupt end-effector orientation changes, we must take special care because it can come into a situation where no one has total control over the end-effector orientation. This is particularly true when robot programs are generated off-line. In some situations this could be a major problem, leading to the appearance of defects in the work produced by the robot. The proposed solution to circumvent this problem is based on the implementation of linear smooth interpolation of end-effector positions and orientations [20]. The process involves the following steps:

1. Identification of risk areas. This is done by analyzing the CAD drawing of the cell and manually defining those areas in the drawing (abrupt end-effector orientation changes);

2. Discretization of the risk robot path in equally spaced intervals;

3. Calculation of end-effector orientations for each interpolated path point. The new path is smoother than the initial, Fig. 15.

**Fig. 15.** End-effector poses before interpolation (a) and after interpolation (b).

For the profile in Fig. 15, interpolation was divided in two sections $S_1 \in [\mathbf{P}_j, \mathbf{P}_{j+1}]$ and $S_2 \in [\mathbf{P}_{j+1}, \mathbf{P}_{j+2}]$. The calculations are presented for section $S_1$ but for other sections the procedure is the same. For a sampling width $\Delta t$ the interpolated position $\mathbf{r}(k) = (r_x, r_y, r_z)^T$ is:

$$r_i(k) = r_i(0) + v_i(k)\, k\, \Delta t, \quad \begin{cases} (i = x, y, z) \\ k = 1, \dots, n-1 \end{cases} \tag{14}$$

Where $v_i(k)$ is a directional velocity profile. Note that *n* represents the number of interpolated points.

A spherical linear interpolation (SLERP) algorithm was implemented for the purpose of quaternion interpolation. Given two known unit quaternions, $\mathbf{Q}_0$ (from $\mathbf{P}_j$) and $\mathbf{Q}_n$ (from $\mathbf{P}_{j+1}$) with parameter *k* moving from 1 to *n*-1, the interpolated end-effector orientation $\mathbf{Q}_k$ can be obtained as follows:

$$\mathbf{Q}_k = \frac{\sin\left(\left(1 - \frac{k-1}{n-1}\right)\theta\right)}{\sin\theta} \mathbf{Q}_0 + \frac{\sin\left(\frac{k-1}{n-1}\theta\right)}{\sin\theta} \mathbf{Q}_n, \quad k = 1, \dots, n-1 \tag{15}$$

Where:

$$\theta = \cos^{-1}(\mathbf{Q}_0 \cdot \mathbf{Q}_n) \tag{16}$$

This method for quaternion interpolation is also used when we want to interpolate Euler angles, simply by transforming Euler angles into quaternions and vice versa.

*3.7. Generation of robot programs*

The demand for intuitive ways to programme machines has led to the emergence of techniques to generate machine code. In the last few years, several code generation techniques have been developed, for example, today's commercial CAD/CAM systems are able to generate cutter location data for CNC machining [31] or to generate CNC tool paths from standard CAD formants [32]-[34]. Nevertheless, these systems tend to have drawbacks such as their ability to generalize from different situations and respond to unseen problems. During the elaboration of an algorithm to generate code, the keyword is "generalize" and never "particularize", the algorithm must be prepared to cover a wide range of variations in the process. For particular applications with a limited and well known number of process variations this kind of systems tend to present good performance [35]. In fact, today, reliable CNC programs are often generated from CAM software.

Robot controller specific languages have seen only minor advances in the last few years. Some authors have devoted attention to create methodologies capable to generalize robot programs around a task but which at the same time can be customized as necessary [36]. For example, an operation can be customized in terms of type of robot operation or shape of the product in production. Intrinsically, this is a way to profit from previously similar work, incorporating the programmers' experience and process knowledge [37]. In this way, the time to produce programs for related products/tasks can be reduced and non-specialists can create robot programs by themselves. These systems follow the same logic as the well known macros or scripts in the world of computer science.

In the context of this paper, we propose an algorithm to automatically generate robot programs with information extracted from CAD drawings. The way the process to generate robot code is applied differs from situation to situation. Nevertheless, there is a common point in all robot programs. It means that since robots usually perform manipulation tasks, the process to generate a robot program does not differ greatly from application to application, containing common tasks like gripping, moving and placing, Fig. 16**Error! Reference source not found.**. The proposed system was developed having in mind that the generated robot programs should be generalist, feasible and reliable as possible.

**Fig. 16.** The different phases of a manipulation task.

The generation of a robot program is no more that writing robot commands in a text file, line by line. This process is managed by the software interface that extracts data from CAD drawings, interprets that data and finally generates robot programs. Some robot and process parameters are defined in the software

interface by the user. The proposed process to generate a robot program is divided into two distinct phases:

1. Definition and parameterization of robot end-effector poses (positions and orientations), frames, tools and constants. The algorithm in Fig. 17 summarizes the process of data acquisition from tool models and/or path lines, and the generation of robot code. The following equation represents a common definition of a robot pose.

$$P = x, y, z, \begin{cases} \alpha, \beta, \gamma & \longrightarrow \text{Euler angles} \\ q_w, q_x, q_y, q_z & \longrightarrow \text{Quaternions} \end{cases} \quad (17)$$

   When confronted with risk areas, the interpolation algorithms automatically generate the appropriate end-effector positions and orientations for those areas. Specific process and robot parameters (coordinate systems, tools, etc.) are specified in this phase, as well as the necessary constants. Usually, this information comes from the parameters introduced in the software interface, for example, robot home position, number of working cycles, approaching distances, etc.;

2. Body of the program. A robot program contains predominantly robot motion instructions, linear, joint, circular or spline robot movement. These movement instructions are selected according to the type of motion established by the user in the CAD drawing and/or the software interface. For example, if a segment of a path is drawn as a straight line, the generated code will contain a robot instruction that makes the robot end-effector move linearly in that path segment. In this phase the algorithm also deals with particular situations associated with each robotic task such as IO commands to communicate with other machines and the definition of safety areas. An important issue has to do with the openness of the proposed architecture which has an open interface to extend its functionality and by this way be adaptable for specific robot tasks.

The proposed algorithm is able to generate robot programs for *Motoman* (Fig. 18) and *ABB* robot controllers, INFORM and RAPID languages respectively. However, as all robot programs are based on the same principle, the proposed algorithm can be adapted to generate code in other programming languages.

**Fig. 17.** The process of extracting data from CAD and the generation of robot code.

**Fig. 18.** A snippet of a robot program generated for a *Motoman* robot.

*3.8. Software interface*

The software interface makes the link between the user, *Autodesk Inventor* and the robot. Two major requirements for the software interface are simplicity and user friendliness. The functionalities and architecture of the proposed software interface is schematically shown in Fig. 19. This software interface (Fig. 20) runs under *Microsoft Windows* operating systems (*XP* or above) and in any industrial or personal computers with processing and graphical capacity to host *Autodesk Inventor*. The software was mainly written in VB.

**Fig. 19.** Functionalities and architecture.

**Fig. 20.** Graphical user interface.

**4. Experiments**

The CAD-based OLP solution presented in this paper was validated on two different experimental setups. The first involves a robot manipulating objects from a location to another and the second a robot transporting an object between obstacles.

*4.1. Experiment I*

This experimental setup was designed to accomplish a simple object manipulation task. Robot programs are generated from a CAD *assembly model* of the cell in study, Fig. 21, where the robot tool models represent the target poses for robot motion. The robot task from which a robot program is generated consists in having the robot handling three objects from an initial to a final pose, Fig. 22 [38].

**Fig. 21.** Two different perspectives of the CAD *assembly model*.

**Fig. 22.** Robot running the program generated from CAD.

*4.2. Experiment II*

In this experiment, robot programs are generated from a CAD *assembly model* of the cell in which the virtual paths (positional data) are represented in the form of straight lines, arcs and splines. Robot end-effector orientation is defined by placing tool models along the above mentioned paths. These models define the orientation of the robot end-effector in each segment of the path, Fig. 23. The robot program generated from CAD is tested in a real scenario. As shown in Fig. 24 the real robot performs the experienced manipulation task with success bypassing the obstacles without hitting them [39].

**Fig. 23.** CAD *assembly model* of the system in study: with obstacles in invisible mode (a) and in visible mode (b).

**Fig. 24.** Robot running the program generated from CAD.

*4.3. Results and discussion*

The experiments demonstrated the versatility of the proposed CAD-based OLP system. Robotic cell design and robot programming are embedded in the same interface and work through the same platform, *Autodesk Inventor*, without compatibility issues. In terms of accuracy, as in the case of commercial OLP software, the error that may exist comes from inaccuracies in the calibration process and/or in the construction of the CAD models. In fact, error is always present in a calibration process, which may or may not be acceptable, depending on their magnitude and application under consideration. Often, calibration errors arise from the little time and attention devoted to the calibration process (accuracy generally increases proportionally with the time devoted to the calibration process). This situation is increasingly common as companies are constantly being asked to change production. The above is true for all the robot programming and simulation systems based on virtual representation of objects.

The proposed CAD-based HRI system is not the definitive solution for OLP. Nevertheless, it is an original contribution to the field, with pros and cons. The proposed system is limited in some aspects, for example in the level of sophistication and ability to generalize from particular situations. In the other hand, the intuitiveness of use, short learning curve and the low-cost nature of the system appear as positive aspects. All of these characteristics are fundamental when the objective is to spread the utilization of this kind of systems in SMEs or use it for educational and training purposes.

Finally, it is also important to note that after generating a robot program, it should be simulated in order to better visualize the robotic process (robot motion, possible collisions, the re-grasping operations, kinematic singularities, robot joint limits). Since many industrial robots have six degrees of freedom (DOF) and some robotic applications can be performed with five DOF, the reminding DOF provides "space" to deal with kinematic singularities.

**5. Conclusions and contributions**

A novel CAD-based OLP platform has been studied, developed and tested. Robotic cell design and OLP are embedded in the same interface and work through the same platform, a common commercial CAD package, *Autodesk Inventor*. It was presented a method to extract robot paths (positions and orientations) from a CAD drawing of a given robotic cell. Such data are then treated and transformed into robot programs. In addition, experiments showed that the proposed system is intuitive to use and has a short learning curve, allowing non-experts in robotics to create robot programmes in just few minutes. In terms of accuracy, the error that may exist in the processes of OLP comes from inaccuracies in the calibration process and from situations where the CAD models do not reproduce properly the real scenario.

*5.1. Future work*

There are some aspects of the proposed CAD-based solution that should be improved in future. One aspect has to do with the algorithm to generate code, it has to be more generalist, flexible and easier to tune. An idea for future work is to have a graphical- or icon-based interface to tune the algorithm to generate code in more intuitive way. The other aspect has to do with the robot calibration process. External sensing can help to deal with this situation by increasing the accuracy of the processes, making it less susceptible to error and simpler.

**Figure 1**

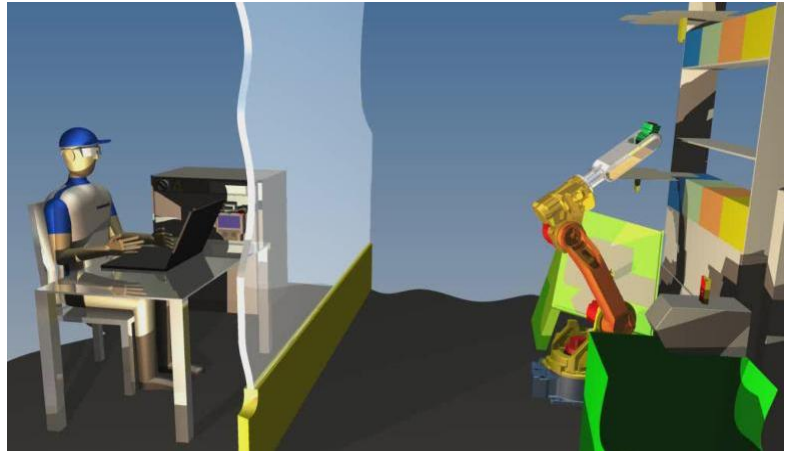

**Fig. 1.** OLP concept.

**Figure 2**

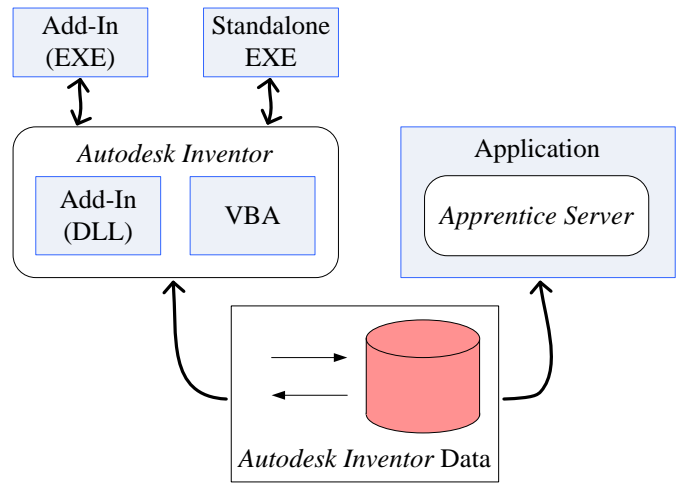

**Fig. 2.** Accessing the Autodesk Inventor's API.

**Algorithm 1. Opening *Autodesk Inventor* (coded in VB).**

```vb
1  ' Get object.
2  Private oApp As Inventor.Application
3  Try
4      oApp =
         System.Runtime.InteropServices.Marshal.GetActiveObject("Inventor.Application")
5  Catch ex As Exception
6      MessageBox.Show("A problem occurs")
7  End Try
8  ' Open Inventor (oApp).
9  Dim InventorAppType As Type = System.Type.GetTypeFromProgID("Inventor.Application")
10 oApp = System.Activator.CreateInstance(InventorAppType)
11 ' Make Inventor visible.
12 oApp.Visible = True
```

**Fig. 3.** Opening *Autodesk Inventor* (coded in VB).

**Figure 4**

```vb
1   ' Open an Inventor document (InventorDoc).
2   Private oApp As Inventor.Application
3   Private InventorDoc As Inventor.Document
4   InventorDoc = oApp.Documents.Open("document name")
5   ' Properties of the document.
6   Dim oPropsets As PropertySets
7   oPropsets = InventorDoc.PropertySets
8   ' Name of the document.
9   Dim doc_name As String
10  doc_name = InventorDoc.DisplayName
```

**Algorithm 2.** Opening an *Autodesk Inventor* document (coded in VB).

**Fig. 4.** Opening an *Inventor* document and extracting its properties and name (coded in VB).

**Figure 5**

**Algorithm 3. Extracting data from a selected *WorkPoint* (coded in VB).**

```vb
1   ' Select an item.
2   Dim oSelectSet As SelectSet
3   oSelectSet = ThisApplication.ActiveDocument.SelectSet
4   ' Check if the selected item is a WorkPoint.
5   If TypeOf oSelectSet.Item(i) Is WorkPoint Then
6       Dim wp1(i) As WorkPoint
7       wp1(i) = oSelectSet.Item(i)
8       ' WorkPoint data (name, X, Y, Z).
9       WorkPointPos(i).oName = wp1(i).Name
10      WorkPointPos(i).x = wp1(i).Point.X
11      WorkPointPos(i).y = wp1(i).Point.Y
12      WorkPointPos(i).z = wp1(i).Point.Z
13  Else
14      MsgBox("You must select a WorkPoint.")
15      Exit Sub
16  End If
```

**Fig. 5.** Extracting data from a selected *WorkPoint* (coded in VB).

**Algorithm 4. Extracting data from a selected virtual line (coded in VB).**

```vb
1   'Extracting straight line data
2   If TypeOf oSelectSet.Item(i) Is SketchLine3D Then
3       'Defining an object type SketchLine3D
4       Dim Line_ As SketchLine3D
5       Line_ = oSelectSet.Item(i)
6       'Start and end points of the SketchLine3D
7       Dim start_point_x, start_point_y, start_point_z As Double
8       Dim end_point_x, end_point_y, end_point_z As Double
9       start_point_x = Line_.StartSketchPoint.Geometry.X
10      start_point_y = Line_.StartSketchPoint.Geometry.Y
11      start_point_z = Line_.StartSketchPoint.Geometry.Z
12      end_point_x = Line_.EndSketchPoint.Geometry.X
13      end_point_y = Line_.EndSketchPoint.Geometry.Y
14      end_point_z = Line_.EndSketchPoint.Geometry.Z
15      'Extracting spline data
16  ElseIf TypeOf oSelectSet.Item(i) Is SketchSpline3D Then
17      'Defining an object type SketchSpline3D
18      Dim Spline_ As SketchSpline3D
19      Spline_ = oSelectSet.Item(i)
20      'Start, medium and end points of the SketchSpline3D
21      Dim s_start_point_x, s_start_point_y, s_start_point_z As Double
22      Dim s_mid_point_x, s_mid_point_y, s_mid_point_z As Double
23      Dim s_end_point_x, s_end_point_y, s_end_point_z As Double
24      s_start_point_x = Spline_.StartSketchPoint.Geometry.X
25      s_start_point_y = Spline_.StartSketchPoint.Geometry.Y
26      s_start_point_z = Spline_.StartSketchPoint.Geometry.Z
27      s_mid_point_x = Spline_.FitPoint(2).Geometry.X
28      s_mid_point_y = Spline_.FitPoint(2).Geometry.Y
29      s_mid_point_z = Spline_.FitPoint(2).Geometry.Z
30      s_end_point_x = Spline_.EndSketchPoint.Geometry.X
31      s_end_point_y = Spline_.EndSketchPoint.Geometry.Y
32      s_end_point_z = Spline_.EndSketchPoint.Geometry.Z
33  End If
```

**Fig. 6.** Extracting data from a selected virtual line (coded in VB).

**Figure 7**

```vb
Algorithm 5. Transformation matrix (coded in VB).
1   ' Get an occurence from the selected item.
2   Dim oOccurrence As ComponentOccurrence
3   oOccurrence = ThisApplication.ActiveDocument.SelectSet.Item(i)
4   ' Get the transformation matrix.
5   Dim oTransform As Inventor.Matrix
6   oTransform = oOccurrence.Transformation
7   ' Get matrix data, for example cell (1, 3).
8   Dim mt13 As Double
9   mt13 = oTransform.Cell(1, 3)
```

**Fig. 7.** Extracting the transformation matrix of a selected item (coded in VB).

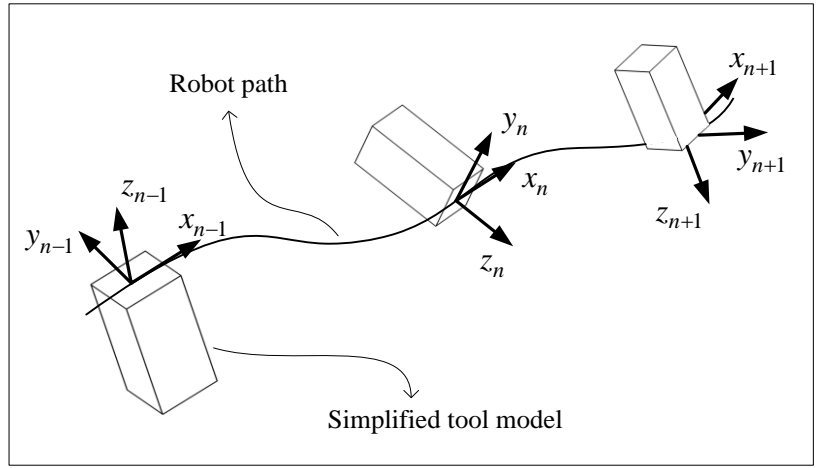

**Fig. 8.** Simplified tool models defining the robot end-effector pose in each path segment.

**Figure 9**

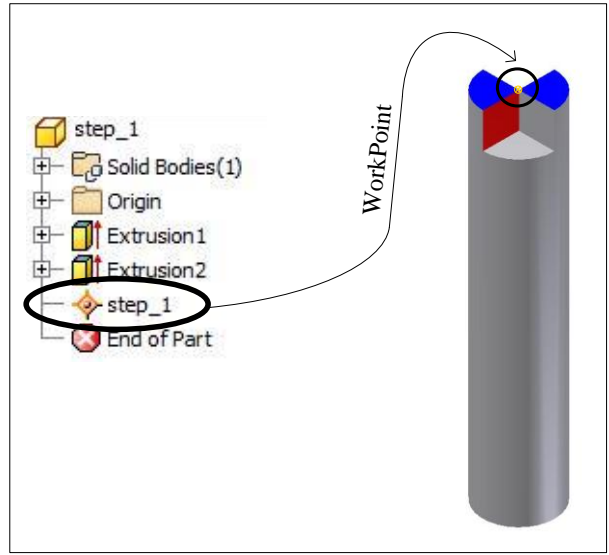

**Fig. 9.** A *WorkPoint* attached to a tool model in a location where the tool is connected to the robot wrist.

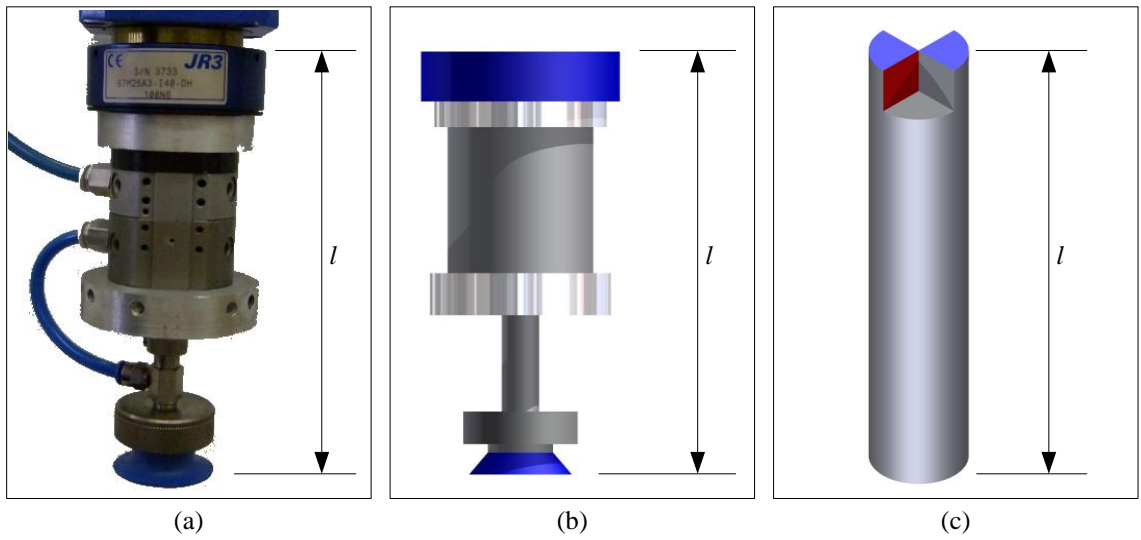

**Fig. 10.** Real robot tool (a), and simplified models (b) and (c).

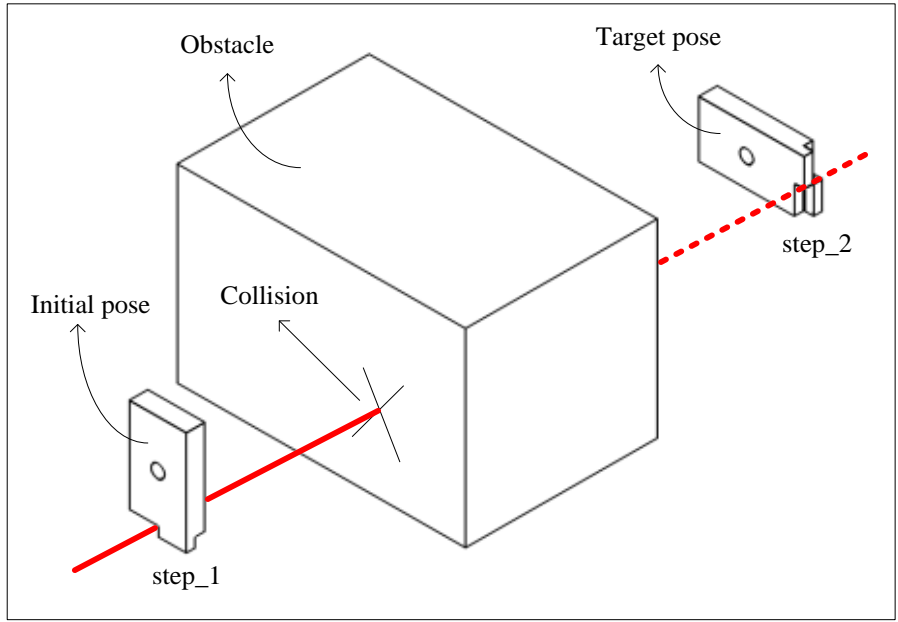

**Fig. 11.** Collision.

**Figure 12**

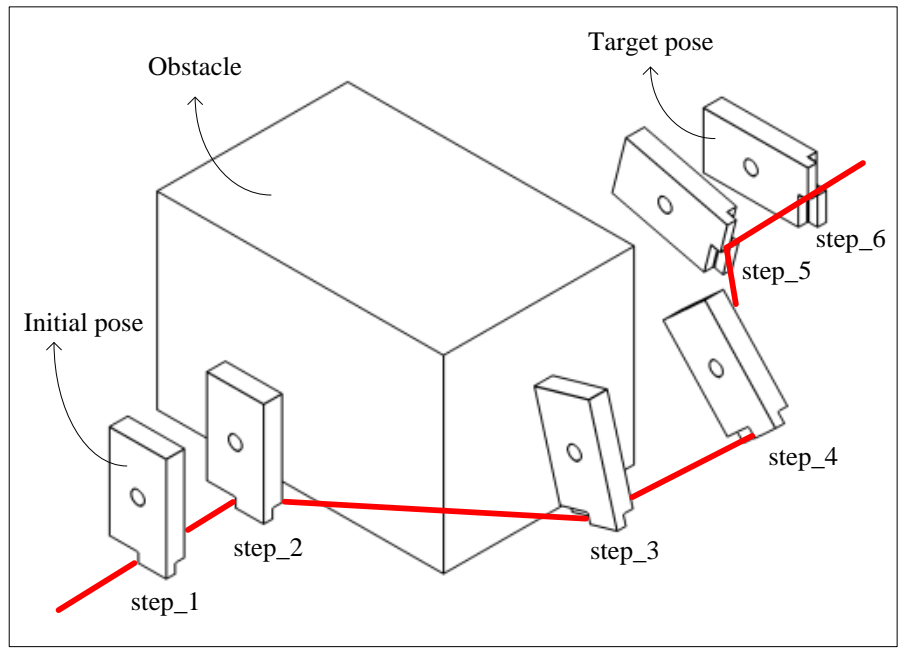

**Fig. 12.** Avoiding an obstacle.

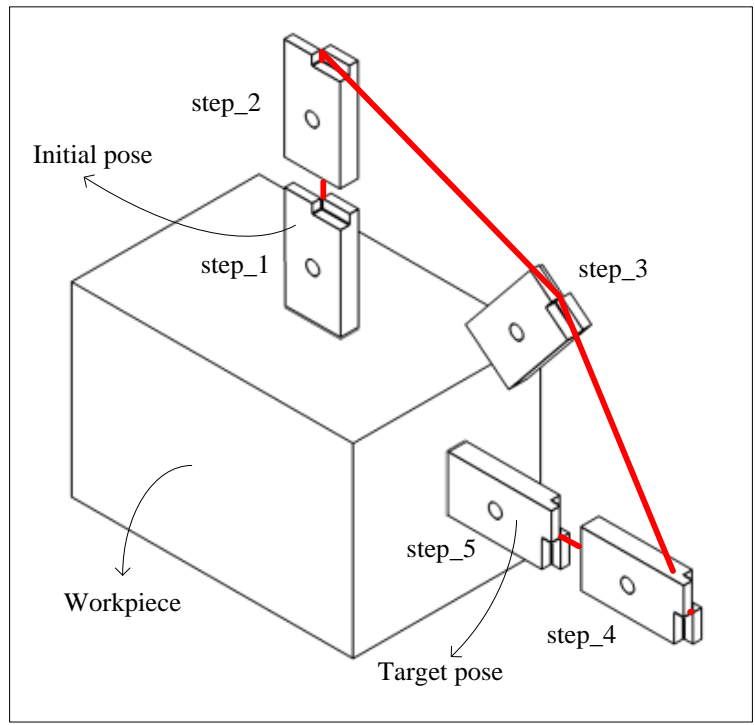

**Fig. 13.** Re-grasping.

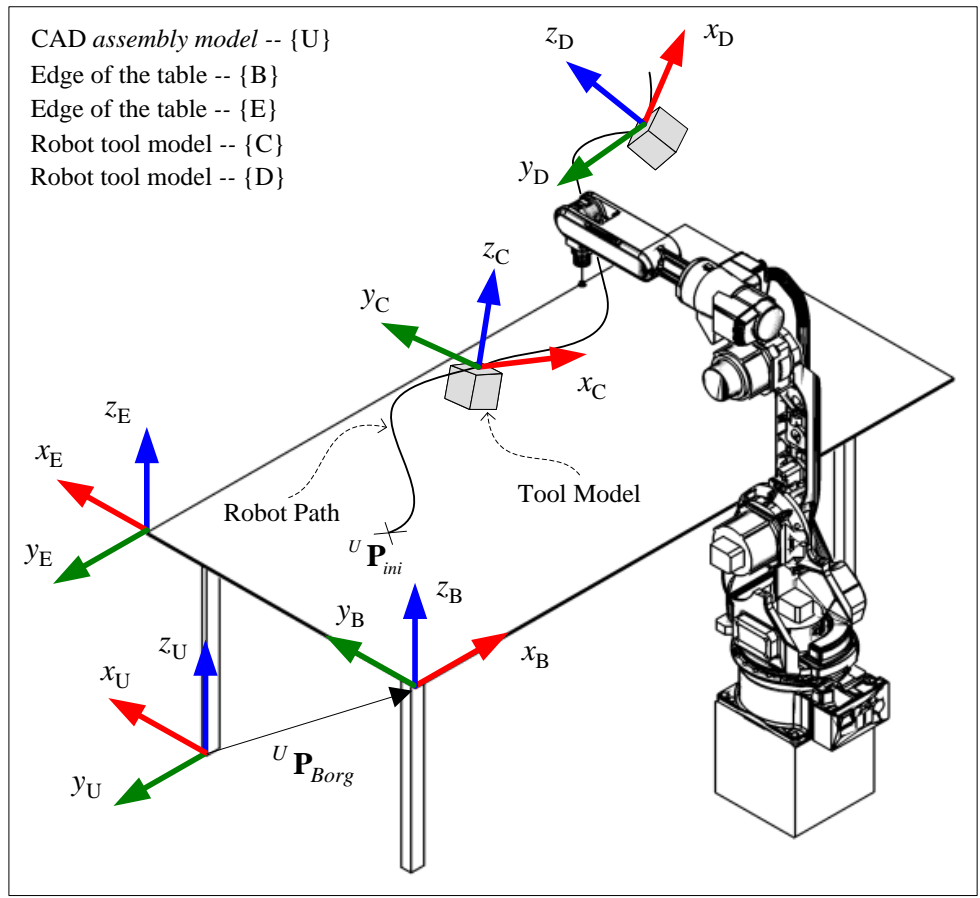

**Fig. 14.** Coordinate frames.

**Figure 15**

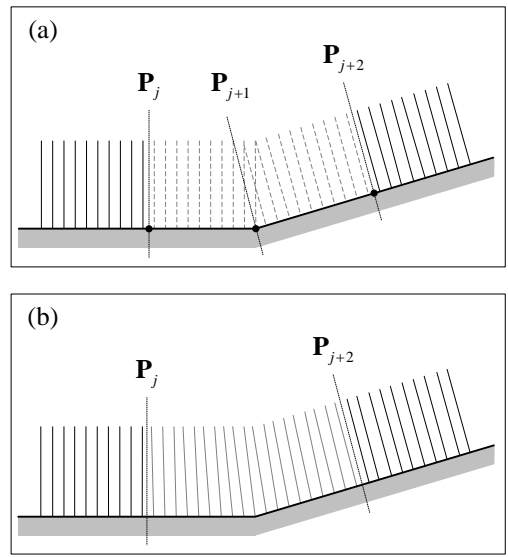

**Fig. 15.** End-effector poses before interpolation (a) and after interpolation (b).

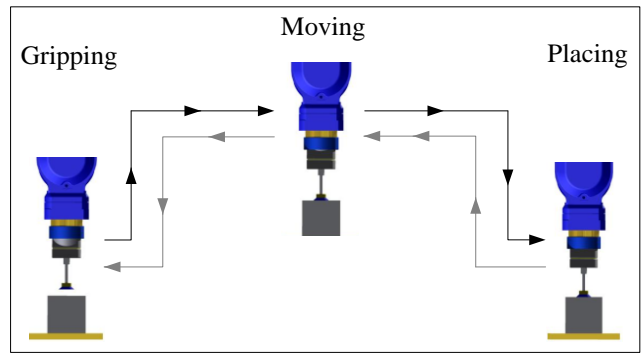

**Fig. 16.** The different phases of a manipulation task.

```
1   For Each tool model
2       Get WorkPoint position
3       Get transformation matrix
4       Calculate frame correlations
5       Calculate Euler angles
6       If (tool name = step_n)
7           Generate code for a simple robot movement
8       Else If (tool name = step_n…)
9           Generate code for a "special" robot operation
10      End If
11  End
12  For Each selected path line
13      Get path position
14      Get end-effector orientation from a tool model
15      Calculate frame correlations
16      Generate code for a simple robot movement
17  End
```

**Fig. 17.** The process of extracting data from CAD and the generation of robot code.

```
P00050=336.99820026424,-149.99999...
P00053=336.94411482772,-144.99998430324,670,0,0,0
P00064=489.3844547939,403.24359985482,775.050253609301,0,0,0
P00070=489.38445479391,166.36209915648,775.050253609301,0,0,0
//INST
///DATE 2010/06/29 16:28
///ATTR SC,RW
///GROUP1 RB1
NOP
SET D001 0
SET D002 1
SET I005 1
SET I000 10
*LABEL1
DOUT OT#(1) OFF
MOVJ P0033 VJ=23.00
MOVJ P001 VJ=23.00
JUMP *LABEL2 IF I000=10
IMOV P0036 V=23.00 TF
*LABEL2
ADD P0036 P0043
DOUT OT#(1) ON
TIMER T=1.00
MOVJ P0033 VJ=23.00
```

**Fig. 18.** A snippet of a robot program generated for a *Motoman* robot.

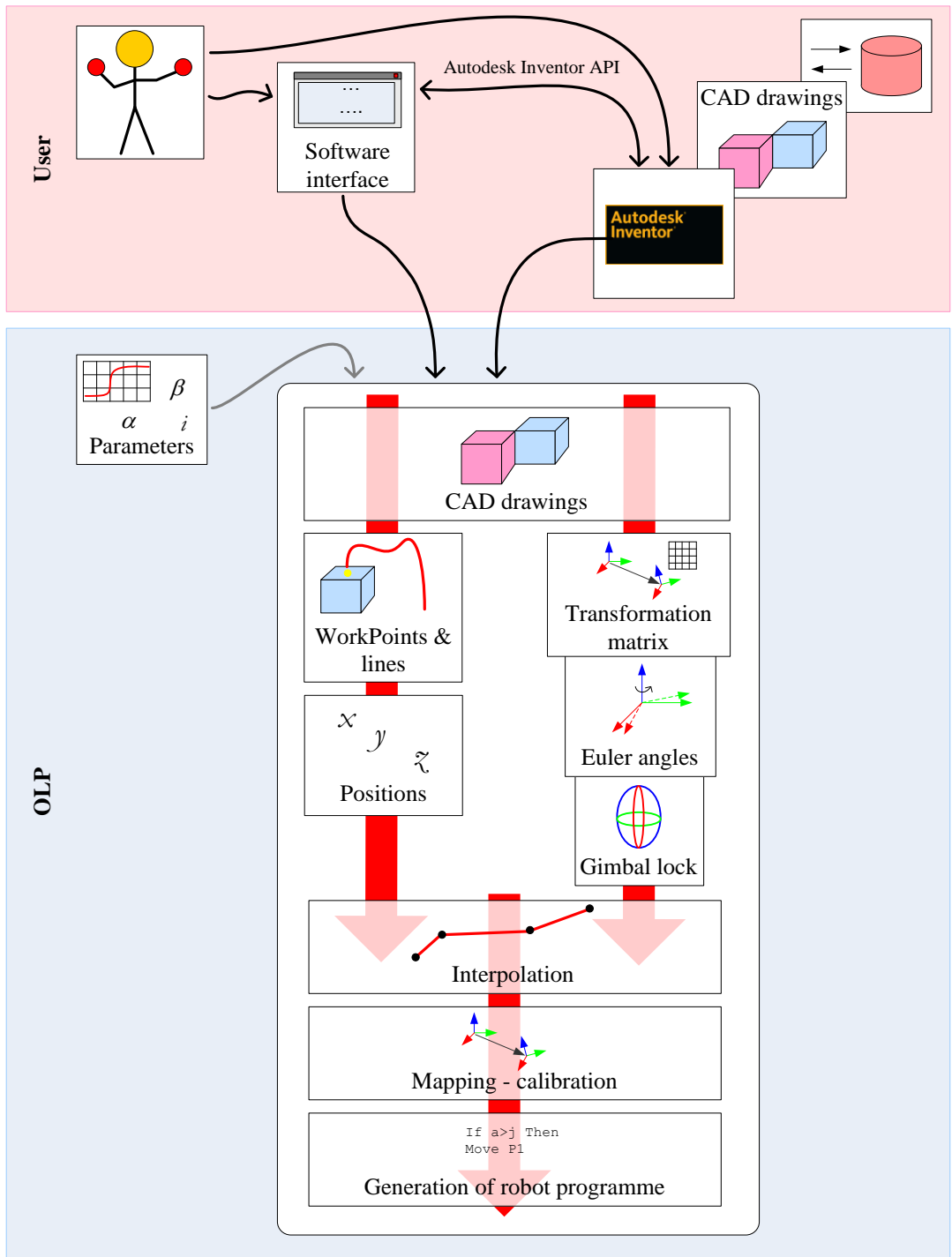

**Fig. 19.** Functionalities and architecture.

**Figure 20**

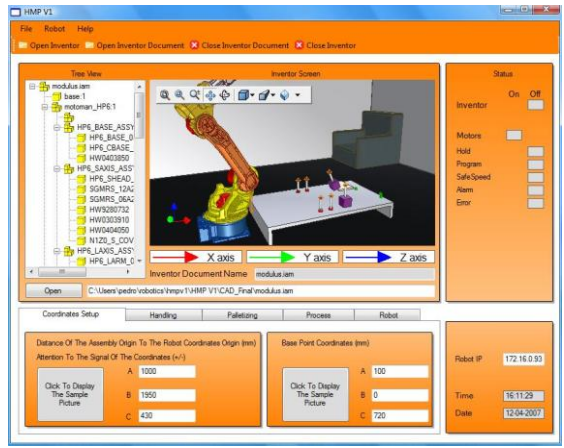

**Fig. 20.** Graphical user interface.

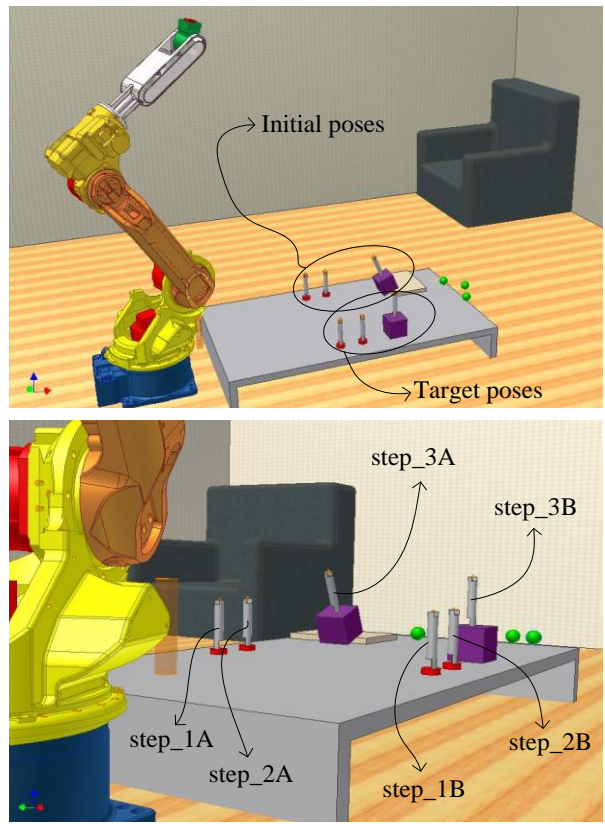

**Fig. 21.** Two different perspectives of the CAD *assembly model*.

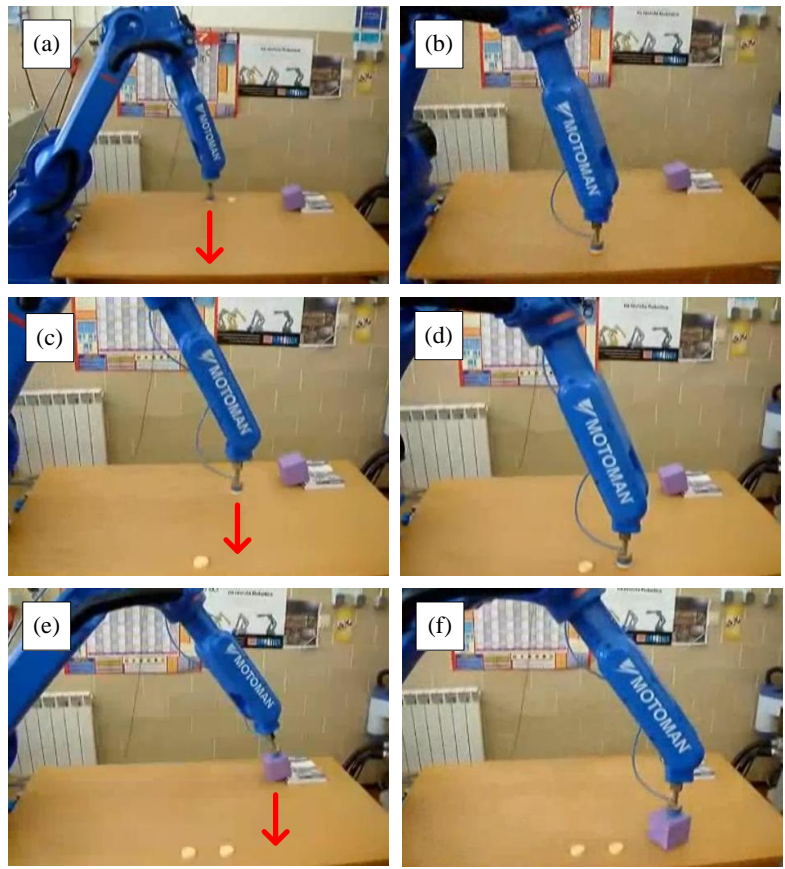

**Fig. 22.** Robot running the program generated from CAD.

**Figure 23**

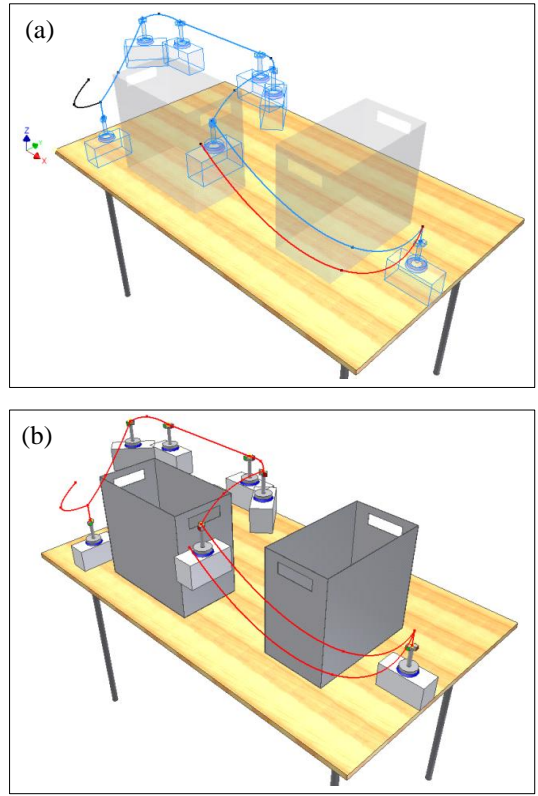

**Fig. 23.** CAD *assembly model* of the system in study: with obstacles in invisible mode (a) and in visible mode (b).

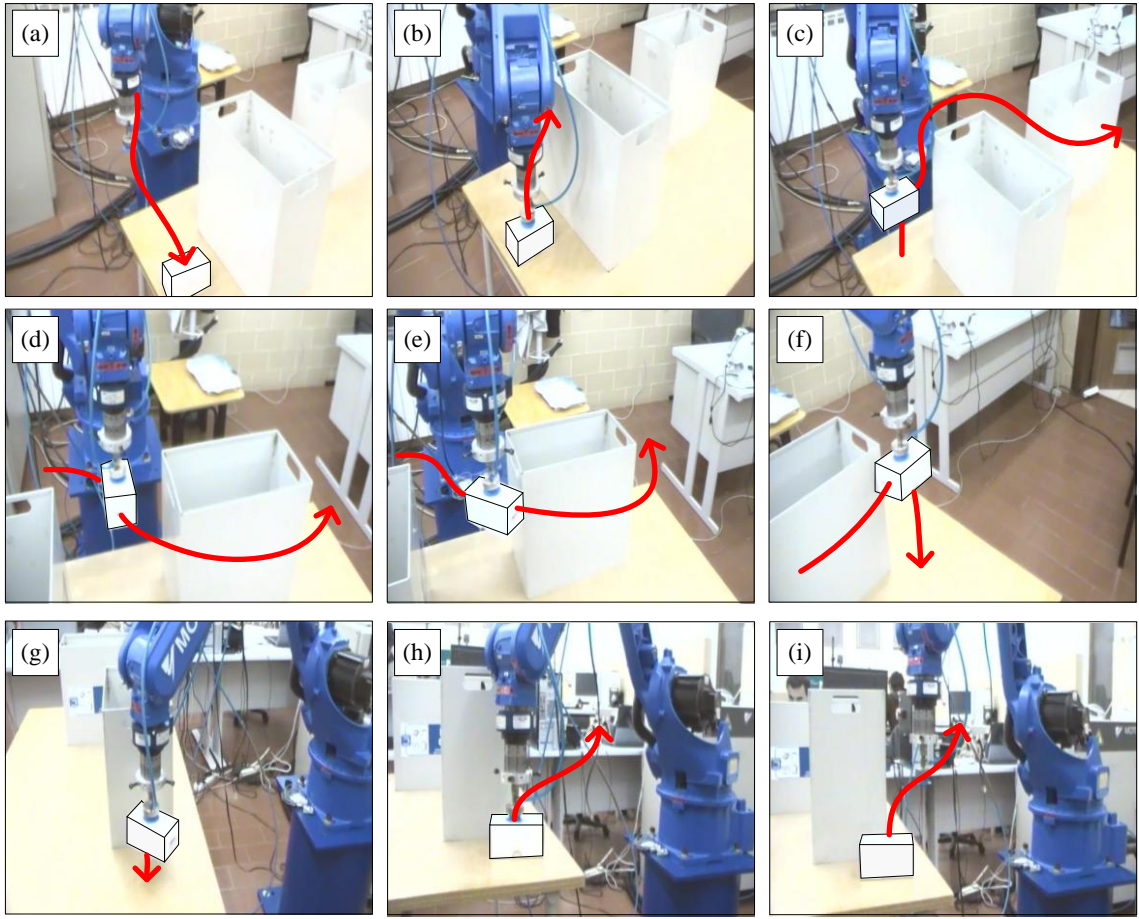

**Fig. 24.** Robot running the program generated from CAD.

**Biography Pedro Neto**

Pedro Neto was born in Coimbra, Portugal, on February 1, 1984. He received the Bachelor degree and Ph.D. degree in Mechanical Engineering from the University of Coimbra in 2007 and 2012, respectively. He has been involved in teaching activities in Automation and Industrial Control since 2010 as Assistant Professor at the Department of Mechanical Engineering of the University of Coimbra. His research interests include: human-robot interaction, pattern recognition, CAD-based robotics and sensor fusion. Pedro Neto is author of several journal and conference publications. He participated in two European funded R&D projects, FP6 and FP7 and national projects.

*Biography of Nuno Mendes

**Biography Nuno Mendes**

Nuno Mendes is currently a Ph.D. student at the University of Coimbra. He received the Bachelor degree in Mechanical Engineering from the University of Coimbra in 2008. His research interests include: CAD-based robotics, sensor fusion, force control, Fuzzy and robotic friction stir welding. Nuno Mendes is author of several journal and conference publications.

*Photo of Pedro Neto

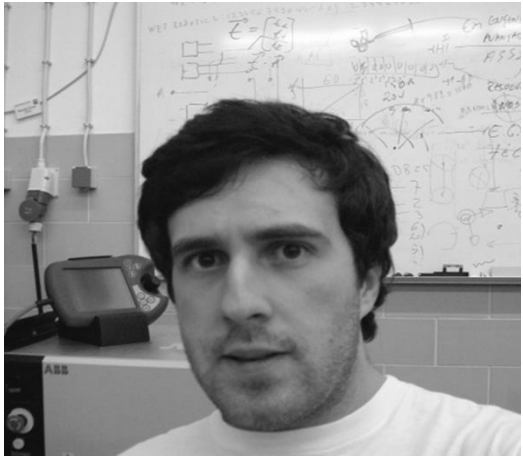

**Photo of Nuno Mendes**

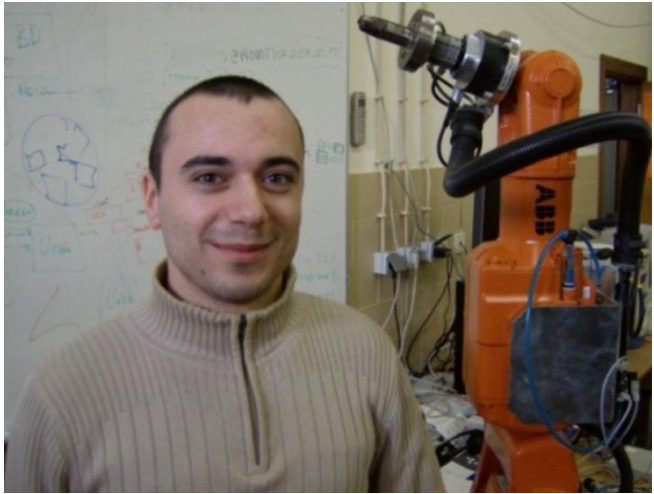